\pdfoutput=1

\documentclass[11pt]{article}

\usepackage[final]{acl}

\usepackage{times}
\usepackage{latexsym}
\usepackage{amsmath}
\usepackage{kotex}
\usepackage{adjustbox}
\usepackage{float}
\usepackage{dirtytalk}
\usepackage{algorithm}
\usepackage{algorithmic}
\usepackage[T1]{fontenc}

\usepackage[utf8]{inputenc}

\usepackage{microtype}

\usepackage{inconsolata}

\usepackage{graphicx}

\newcommand{\psl}{PSL}
\newcommand{\pslfull}{Probabilistic Soft Logic (PSL)}
\newcommand{\sref}[1]{\S\ref{#1}} %

\usepackage{graphicx}               
\usepackage{tabularx}               
\usepackage{booktabs}
\usepackage{amsmath}
\usepackage{stmaryrd}
\usepackage{xcolor}
\usepackage{soul}

\newcolumntype{C}{>{\centering\arraybackslash}X}
\usepackage{multirow}               
\usepackage{diagbox}                
\usepackage{hhline}                 
\usepackage{color}                  
\usepackage{amsmath}                
\usepackage{amssymb}                
\usepackage{mathtools}              
\usepackage{enumitem}               

\usepackage{subfigure}
\usepackage{booktabs}
\usepackage{extarrows}
\usepackage{makecell}

\usepackage{soul}

\definecolor{RoseQuartzBg}{HTML}{F7CAC9}
\definecolor{RoseQuartz}{HTML}{F5A798}
\definecolor{Serenity}{HTML}{92A8D1}
\definecolor{OrangeRed}{rgb}{1.0, 0.27, 0.0}
\definecolor{Red}{rgb}{1.0, 0.0, 0.0}
\definecolor{Turquoise}{HTML}{0F4C81}
\usepackage{xparse}
\NewDocumentCommand{\js}{ mO{} }{\textcolor{OrangeRed}{\textsuperscript{\textit{Jong}}\textsf{\textbf{\small[#1]}}}}
\NewDocumentCommand{\nishant}{ mO{} }{\textcolor{blue}{\textsuperscript{\textit{Nishant}}\textsf{\textbf{\small[#1]}}}}
\NewDocumentCommand{\wenlong}{ mO{} }{\textcolor{Serenity}{\textsuperscript{\textit{Wenlong}}\textsf{\textbf{\small[#1]}}}}

\NewDocumentCommand{\jy}{ mO{} }{\textcolor{RoseQuartz}{\textsuperscript{\textit{jy}}\textsf{\textbf{\small[#1]}}}}
\NewDocumentCommand{\mh}{ mO{} }{\textcolor{RoseQuartz}{\textsuperscript{\textit{mh}}\textsf{\textbf{\small[#1]}}}}

\title{An Analysis under a Unified Formulation of\\Learning Algorithms with Output Constraints}

\author{Mooho Song \\
  Seoul National University \\
  \texttt{anmh9161@snu.ac.kr} \\\And
  Jay-Yoon Lee \\
  Seoul National University \\
  \texttt{lee.jayyoon@snu.ac.kr} \\}

\begin{document}
\maketitle
\begin{abstract}
Neural networks (NN) excel in diverse tasks but can produce nonsensical results due to their exclusive reliance on (input, output) pairs, often conflicting with human knowledge. Injecting human knowledge via output constraints can enhance performance and reduce violations. Despite attempts to compare existing algorithms, no unified categorization of learning algorithms with output constraints exists.
Our contributions are:
(1) We categorize previous studies using three axes: type of constraint loss (e.g., probabilistic soft logic, REINFORCE), exploration strategy of constraint-violating examples, and integration mechanism for balancing the main task and constraints learning signals.
(2) We propose new algorithms inspired by continual-learning for integrating main task and constraint information.
(3) We introduce the $H\beta$-score metric to simultaneously evaluate main task performance and constraint violation.
Our experiments on NLP tasks (NLI, STE, SRL) show that our projection-based integration mechanism outperforms others. Sampling strategy is crucial for high $H\beta$-scores, with better results as sample numbers increase. Additionally, soft-type constraint loss performs well when combined with sampling strategies. These insights highlight key factors for achieving high $H\beta$-scores and demonstrate the efficacy of our methods.
\end{abstract}

\section{Introduction}
The majority of neural networks (NN) models \say{solely} learn from data in the form of (input, output) pairs, and such models can sometimes result in a conflict with human knowledge. Previous work has shown that injecting human knowledge into NN models in the form of reducing relevant constraint violations during training time can improve the model performance as well as reducing constraint violations \cite{li-etal-2020-structured,nandwani2019primal,mehta2018towards,gluecons,xu2018semantic}.
 The relation between constraint and the task itself can be viewed as a relation between the sub-task and the main task. The goal of the main task would be to acquire the most accurate prediction possible, whereas the goal of the subtask is to simply acquire constraint-satisfying output.
 The focus in injecting constraint is to preserve or improve the main task performance while improving the constraint satisfaction.
 
Various literature exists on constraint injection during training time ~\cite{li-etal-2020-structured,nandwani2019primal,mehta2018towards}, where the majority of them formulates the loss function as an addition of loss related to constraint to the existing supervised loss term. Research on how to formulate the loss related to constraint and how much to incorporate in comparison to existing supervised loss is scattered as these approaches vary across different studies and applications.

The first goal of this work is to provide a unified analysis of existing methods from previous studies under a single mathematical formulation. Early efforts to compare different constraint injection methods \cite{gluecons} do exist, however, their focus was on comparing performances of different algorithms, as presented in previous work.
On the other hand, our study aims to formalize previous literature from a new unified perspective, to understand key success factors in existing algorithms.
For example, while primal-dual algorithm \cite{nandwani2019primal} have shown positive results with the idea of dynamic weight update on constraint-loss, it was only tested under the single loss type of \pslfull\ \cite{brocheler2012probabilistic}. This makes it unclear whether the positive results were contributions mostly coming from \psl\ or from the novel dynamic weight update algorithm.
As the same weight update mechanism can be applied to different loss types, such as REINFORCE loss, it is worthwhile to investigate mixing and matching different components of injecting constraint under a unified formulation.
While numerous studies have focused on injecting constraint during training time, to the best of our knowledge, there has been no research consolidating these studies into a unified mathematical formulation to compare their characteristics component by component. 

The second goal of this work is to propose new effective learning algorithms that integrate constraints within the suggested unified formulation. A common approach to learning with constraints involves handling a constraint loss term, $\lambda \times \mathcal{C}$, where $\mathcal{C}$ denotes the constraint loss and $\lambda$ is 
a fixed scalar
representing the weight of $\mathcal{C}$. By adding $\lambda \times \mathcal{C}$ to the pre-existing supervised loss term. 
\citeauthor{nandwani2019primal} introduced an 
algorithm that dynamically controls $\lambda$, starting training with $\lambda$ at 0 and progressively adjusting its value during the learning process. This algorithm is characterized by the gradual increase of the weight $\lambda$, updating it solely based on the degree of constraints.
While the work of \citeauthor{nandwani2019primal} is distinguished from existing methods that use a fixed hyperparameter $\lambda$, there has not been sufficient research 
for integrating the learning signals form supervised data and constraint information beyond this work.

Is it always necessary to have the value of $\lambda$ monotonically increasing for training? Is there a way to update the value of $\lambda$ considering both supervised learning and constraint injection?
Inspired by continual learning methods, this paper proposes a new approach that considers both supervised loss and constraint loss during gradient updates. This approach takes into account the progress of both tasks: supervised learning and constraint injection tasks.
It offers a new viewpoint for injecting constraint on simultaneously learning these two tasks.
Experiments demonstrate that our new approach achieves the highest-level of performance other learning algorithms in various scenarios.

\section{Unified formulation of previous work} 
\label{sec:Analysis_Axes}
In this section, we 
categorize the previous studies on injecting constraints during training time based on three dimensions: 
type of mathematical expression used for constraint loss (\sref{subsec:Constraint_Loss}),
exploration strategy of constraint-violating examples (\sref{subsec:exploration}), and 
{mechanism for integrating losses from the main task and the constraint injection task} (\sref{subsec:update}). 
A common approach in machine learning is to define a loss function and employ optimization algorithms to update model parameters in the direction of minimizing that loss.
When the labeled data 
$\{(x_{i},y_{i})\}^{N}_{i=1}$ is given, the goal of typical supervised learning is to solve the following optimization problem:
\begin{align} \label{eq:objective_detailed}
\underset{\theta}{\min}\;\cfrac{1}{N}\overset{N}{\underset{i=1}{\sum}}\mathcal{L}(x_{i}, y_{i}; \theta),\;\;\text{\footnotesize or simply}\;\; \underset{\theta}{\min}\;\mathcal{L}(\theta)
\end{align}
, where $\mathcal{L}(x, y; \theta)$ is the standard supervised loss function for the task we are learning. 

Most of the existing constraint injection methods, while differing in specific formulations, inject the constraint information by expanding the loss function in a following manner:
\begin{align}\label{eq:Total_Loss}
\mathcal{T}(\theta)=\lambda_{1}\mathcal{L}(\theta)+  \lambda_{2} \cdot \mathcal{C}(\theta)
\end{align}
, where $\mathcal{C}(\theta)$ is a loss related to the constraint, and $\lambda_{i}$'s are fixed weights, We can further generalize the equation \eqref{eq:Total_Loss} as follow:
\begin{align}\label{eq:Total_Loss_grad}
    \nabla\mathcal{T}(\theta)=\Lambda^{sup}\cdot\nabla\mathcal{L}(\theta)+  \Lambda^{con}\cdot \nabla\mathcal{C}(\theta)
 \end{align}
, where $\Lambda^{sup}$, $\Lambda^{con}$ are usually scalar matrices.
$\mathcal{C}$ reflects the human knowledge the algorithm wants to inject and is typically computed without the true label. To be more precise, for some hard constraints on output labels, $\mathcal{C}(\theta)$ is computed via output $f_{\theta}(x)$ given some unlabeled input $x$. Injecting more than one hard constraint is also possible by expanding $\Lambda^{con}\cdot \nabla\mathcal{C}(\theta)$ to $\overset{K}{\underset{i=1}{\sum}} \Lambda^{con}_{i}\cdot \nabla\mathcal{C}_{i}(\theta)$
in equation \eqref{eq:Total_Loss_grad} , where $K$ is a number of constraints.

To unify and distinguish different algorithms that learn with constraints, we focus on how $\mathcal{C}(\theta)$ is formulated (\sref{subsec:Constraint_Loss}), how constraint-violating examples are explored
 (\sref{subsec:exploration}), and how $\Lambda^{sup}$, $\Lambda^{con}$ (in equation \eqref{eq:Total_Loss_grad}) are determined(\sref{subsec:update}).

\subsection{Type of constraint loss} \label{subsec:Constraint_Loss}
Type of constraint loss is related to how the violation of constraints can be transformed into the form of a differentiable loss function $\mathcal{C}(\theta)$ in equation \eqref{eq:Total_Loss}, which we will refer to as the \textit{constraint loss}. How to convert symbolic constraints into a differentiable loss function can be broadly categorized into two approaches: \pslfull\  and REINFORCE. 
\paragraph{\textbf{\pslfull}}

\psl\ \cite{brocheler2012probabilistic} is associated with expressing logic in terms of probabilities, and research utilizing \psl\ measures the degree of constraint violation in the logic itself, employing it as a loss. 
Gödel, product, Łukasiewicz logics can be primarily used to soften logic \cite{minervini-riedel-2018-adversarially,nandwani2019primal,li-etal-2020-structured}, and these examples are listed in table \ref{table:soft_logic_norms}. Generally, \psl\ is not suitable for representing all types of hard constraints, since it must be converted to linear constraints before they can be directly applied \cite{gluecons}. 
Section \sref{subsec:task_STE} is an example task illustraing the challenges in applying \psl, and the more details are in Appendix \sref{psl_reinforce_comparison}.

\paragraph{\textbf{REINFORCE}}
In contrast, studies employing the REINFORCE \cite{williams1992simple} evaluate whether (or to what degree) the model's output violates constraints. Constraint injection research during training time using REINFORCE 
can be further classified into two ways depending how the reward is formulated.
A simple method is to assign binary reward (e.g.: \{1, 0\}) when the model satisfies or violates the constraints \cite{ahmed2022pylon}. This simple method with binary reward only considers whether the constraint is satisfied or not.
On the other hand, one could make the reward more fine-grained by measuring the degree of constraint violation and assigning real-valued rewards related to it \cite{mehta2018towards}. 
A significant feature of REINFORCE is that the determination of constraint loss relies solely on the rule of assigning rewards based on the presence or absence of constraint violation in sampled examples, regardless of the specific constraint. This differs from \psl\ in that it does not require intricate implementations for generating constraint loss. However, due to the need for sampling procedures, the computational cost is generally higher than when using \psl\ \cite{gluecons}.
\newline

To summarize, \psl\ and REINFORCE are mainly used approach to generate $\mathcal{C}(\theta)$ in eq.\eqref{eq:Total_Loss} to reduce expected constraint violation with following differences. \psl\ defines constraint violation as a continuous measure, while REINFORCE relies on the reinforcement learning paradigm to guide the model towards satisfying constraints. Specifically, the REINFORCE method is divided into two types based on the method of setting rewards: binary rewards and real rewards. More specific comparison between types of constraint losses: \psl\ and REINFORCE is in Appendix \sref{fine_grained_expressiveness}. 

\subsection{Exploration of constraint-violating examples} \label{subsec:exploration}
Let $f_\theta(x)$ represent the output distribution associated with the model $f$ parameterized by $\theta$ given input $x$. 
The identification of constraint-violating examples from $f(x)$ plays a crucial role in determining constraint loss $\mathcal{C}(\theta)$. 
Therefore, exploration of constraint-violating examples can significantly impact the effectiveness and efficiency of constraint learning. The possible questions we have are as follows:
Would it be better to explore the model's approximate output space? Would it be best to examine the model's best possible effort?
Or would it be better to explore by considering all possible probability distributions?
Theses are considered to determine the magnitude of the constraint loss $\mathcal{C}$. For example, in REINFORCE with \{1, 0\} reward, the reward will be 0 if we only visit constraint-violating examples. 

According to the questions posed above, exploration strategies are divided by three, each explained below: \textit{sampling}, \textit{argmax}, and \textit{exhaustive}.

\paragraph{\textbf{Sampling}}
Sampling strategy involves drawing samples from the forward propagation results of the model $f$ to examine different instances that violate constraints. As demonstrated by \cite{ahmed2022pylon}, this method commonly employs the REINFORCE algorithm to incorporate constraint violations into the loss function for the identified examples. The sampling strategy can be applied independently to all combinations for our other analysis axes, specifically concerning the type of constraint loss (\sref{subsec:Constraint_Loss}) and the integration mechanism of learning signals from main task and constraint (\sref{subsec:update}).

\paragraph{\textbf{Argmax (Top-1)}}
Argmax (Top-1) strategy, constraint violation is assesed by choosing the combination with the highest probability from $f(x)$. Following greedy decoding process such as beam search or Vitrerbi decoding \cite{mehta2018towards}, it evaluates constraint violation for the decoded example. Similar to sampling, it evaluates constraint violation for the decoded example, but distinguishes itself by considering the most probable prediction at that moment without multiple samplings. Like the sampling strategy, the argmax strategy can also be applied independently to all combinations 
under our other axes of anlysis.

\paragraph{\textbf{Exhaustive}}
Exhaustive strategy considers 
probabilities of all output class and its combinations.
It is prominently employed in research related to \psl\ \cite{nandwani2019primal,li-etal-2020-structured}. Since there is no sampling involved, it is computationally cost-effective rather than sampling 
strategy. 
When considering the type of constraint loss (\sref{subsec:Constraint_Loss}), our performance evaluation is exclusively conducted using \psl\ for the exhaustive strategy, excluding the REINFORCE in the constraint loss, as it would be impossible to consider all possible combinations in REINFORCE. 
Since the exhaustive strategy can only be applied for constraint loss type of \psl, exhaustive strategy cannot handle all of general type of constraints.
\subsection{Integration mechanism of learning signals from main task and constraint
}  \label{subsec:update}
This section is related to the integration of main task and the constraint information. 
We categorize integration mechanisms of prior studies into \textit{static} and \textit{monotone ($\lambda \uparrow$)}. Additionally, we introduce three new integration mechanisms based on the linear projection: \textit{projection-sup}, \textit{projection-con}, and \textit{projection-both}, which will be discussed in section \sref{sec:PCCW}. We provide detailed explanations of these mechanisms below.
\paragraph{\textbf{Static}}
For constraint loss $\mathcal{C}$, a widely used approach incorporating $\mathcal{C}$ into the existing supervised loss term is to add $\lambda\cdot\mathcal{C}$ to the previous existing loss, where $\lambda$ is a fixed positive real number \cite{ahmed2022pylon,li-etal-2020-structured,mehta2018towards,minervini-riedel-2018-adversarially}. In this approach, the value of $\lambda$ remains unchanged throughout the training process, serving as a constant multiplier that determines the relative influence of $\mathcal{C}$ in comparison to the main task loss $\mathcal{L}$ in eq.\eqref{eq:Total_Loss}.
\paragraph{\textbf{Monotone ($\lambda \uparrow$)}}
On the other hand, the study by \cite{nandwani2019primal} deviates from this by not using a fixed $\lambda$. Instead, it initiates training with $\lambda$ starting from 0 and progressively adjusting its value during the learning process. This concept emerged from the transformation of the constrained optimization problem into a max-min problem, employing alternative updates. In this method, the value of $\lambda$ steadily grows throughout the training, signifying a progressive emphasis on the constraint loss.
\paragraph{\textbf{Projection}}
Unlike previous methods, projection methods perform gradient updates considering the gradients of two losses: $\mathcal{L}$ and $\mathcal{C}$. For both \textit{static} and \textit{monotone ($\lambda \uparrow$)}, $\Lambda$'s are all diagonal matrices in equation \eqref{eq:Total_Loss_grad}. However, the projection method results in non-diagonal matrices depending on the gradients of both loss functions. 
The detailed formulation will be introduced in section \sref{sec:PCCW}.
\newline\newline
It is important to note that the decision on how to integrate two losses ($\mathcal{L}$, $\mathcal{C}$) is entirely separate from the process of formulating the constraint loss $\mathcal{C}$ (\sref{subsec:Constraint_Loss}), and the exploring strategy of constraint-violation examples (\sref{subsec:exploration}). Therefore, adjusting $\Lambda$'s (or, $\lambda$'s) mentioned in this section can be independently combined with other analysis axes. 

\section{Further exploration on integration of main task and constraint information}\label{sec:PCCW}
In this section, we propose new methods for integrating the losses of main task and constraint injection task. Departing from categorized methods used in previous research, `static' and `monotone($\lambda\uparrow$)', we introduce three new integration mechanism for the two losses:
`projection-sup', `projection-con', and `projection-both'.


\paragraph{Motivation}
Gradient Episodic Memory (GEM) \cite{lopez2017gradient} model is designed for continual learning for positive backward transfer, aiming to store memories of previous tasks in such a way that the loss does not increase when learning from new data. It introduces constraints to prevent an increase in loss for previous tasks stored in memory when learning from new data and presents a new minimization problem. A-GEM \cite{chaudhry2018efficient_agem} is a variant of GEM that is designed for effective memory and computational cost, by storing the averaged episodic memory across the all tasks. 
Motivated by these works, we propose a new method for integrating losses -- $\mathcal{L}(\theta)$ and $\mathcal{C}(\theta)$ -- for two tasks. In GEM/A-GEM, whenever new data was deemed to violate positive backward transfer, it applies a projection operation for the gradients to adjust them. We adapt the concept from GEM/A-GEM and utilize it in designing the integration mechanism of main task and constraint information
\paragraph{Method}
Recall that the derivative of loss function with constraint has form of:
\begin{align}
\nonumber
    \nabla\mathcal{T}(\theta)=\Lambda^{sup}\cdot\nabla\mathcal{L}(\theta)+  \Lambda^{con}\cdot \nabla\mathcal{C}(\theta)
\end{align}
Our approach is rooted in the idea that supervised learning and constraint injection are two distinct tasks, and during their respective updates, we can prevent negatively effecting each other by executing a projection of gradient for each other. The followings are explanations of three new algorithms, and the pseudo-codes are available in Appendix \ref{sec:appendix_PCCW_method}.

\textit{Projection-sup} applies the projection method to the gradient of the constraint loss (namely, adjust $\Lambda^{con}$) to prevent it from negatively affecting the supervised learning task, while storing the averaged gradient vector of supervised learning task $g_{sup}$ previously used for training. Mathematically, project $\nabla\mathcal{C}(\theta)$ via:
\begin{equation}\label{eq:agem_sup}
    \text{Proj}(\nabla\mathcal{C}(\theta)) = \nabla\mathcal{C}(\theta) -  \frac{\nabla\mathcal{C}(\theta) \cdot g_{sup}}{g_{sup} \cdot g_{sup}}g_{sup}
\end{equation}
, whenever $\nabla\mathcal{C}(\theta) \cdot g_{sup} < 0$. Then, the vector $\text{Proj}(\nabla\mathcal{C}(\theta))$ satisfies $\text{Proj}(\nabla\mathcal{C}(\theta)) \cdot g_{sup}=0$. This ensures that $\nabla\mathcal{C}$ is transformed orthogonally to $g_{sup}$, preventing it from providing information that contradicts supervised learning.

Conversely, \textit{projection-con} applies the projection method to the gradient of the supervised loss (namely, adjust $\Lambda^{con}$) to prevent it from negatively affecting the constraint injection task, while storing the averaged gradient vector of constraint injection task $g_{sup}$ previously used for training.
Mathematically, project $\nabla\mathcal{L}(\theta)$ via:
\begin{equation}\label{eq:agem_con}
    \text{Proj}(\nabla\mathcal{L}(\theta)) = \nabla\mathcal{L}(\theta) -  \frac{\nabla\mathcal{L}(\theta) \cdot g_{con}}{g_{con} \cdot g_{con}}g_{con}
\end{equation}
, whenever $\nabla\mathcal{L}(\theta) \cdot g_{con} < 0$. Then, the vector $\text{Proj}(\nabla\mathcal{L}(\theta))$ satisfies $\text{Proj}(\nabla\mathcal{L}(\theta)) \cdot g_{con}=0$. This ensures that $\nabla\mathcal{L}$ is transformed orthogonally to $g_{con}$, preventing it from providing information that contradicts constraint injection.

\textit{Projection-both} combines both projection-sup and projection-con, applying projection to both gradients (namely, adjust both $\Lambda^{sup}$ and $\Lambda^{con}$) to ensure that neither task negatively impacts the other. It stores two types of gradients separately by each task used for training before, and apply two projections \eqref{eq:agem_sup} and \eqref{eq:agem_con} together.

\section{Tasks}
\label{sec:selected_tasks}
In this section, we introduce the tasks for which we conduct experiments: Natural Language Inference (NLI), Synthetic Transduction Example (STE), and Semantic Role Labeling (SRL). Additional details about tasks and implementations are explained in Appendix \sref{sec:selected_tasks_app}.

\subsection{Natural Language Inference (NLI)}
\label{subsec:task_NLI}
NLI is a task that involves understanding the logical relationships between pairs of text. Given a premise (P) and a hypothesis (H), the task is to determine whether P entails H, contradicts H, or maintains a neutral relationship with H.
There exists constraints such as if P entails H, then H must not contradict P. We used the five constraints listed in \cite{minervini-riedel-2018-adversarially}, as shown in table \ref{table:NLI Rules} .
The dataset used is SNLI \cite{bowman-etal-2015-large_snli}. 
\subsection{Synthetic Transduction Example (STE)}
\label{subsec:task_STE}
We also present an artificial task utilized in \cite{lee2019gradient}. A sequence transducer \(T:\mathcal{L}_{S} \rightarrow \mathcal{L}_{T}\) converts the source language \(\mathcal{L}_{S}=(az | bz)^{*}\) to the target language \(\mathcal{L}_{T}=(za | bbb)^{*}\), for example, \(T(azbzbz)=zabbbbbb\). 
The constraint imposed involves the relationship between the number of `b' in the source and the target. Specifically, the count of `b' in the target must be exactly three times that in the source. 

\subsection{Semantic Role Labeling (SRL)}
\label{subsec:task_SRL}

SRL is a natural language processing task that predicts the semantic roles of each word in a sentence with respect to a given verb or predicate. The method of our work employed for this purpose is BIO tagging. 

The Unique Core Roles constraint from \cite{li-etal-2020-structured} is applied as a constraint, which means that there can be no more than one occurrence of each core argument. 
For a predicate $u$, if the model predicts the $i$-th word as \texttt{B-X}, then other words in the same prediction should not be predicted as \texttt{B-X}. This can be expressed as follow.
\begin{align}
\nonumber \forall\; u, i \in s, \texttt{X} &\in \mathcal{A}_{core},\\
B_{\texttt{X}}(u, i) &\rightarrow \underset{j \in s, j \neq i}{\bigwedge} \neg B_{\texttt{X}}(u, j).
\end{align}
The dataset we used is English Ontonotes v5, with the CoNLL-2012 shared task format \cite{pradhan2012conll}. 

\section{Experiments}\label{sec:experiments}
Our experiment is composed of NLI, STE, and SRL tasks, with accuracy, token accuracy, and F1 score are used as the main task metrics, respectively. 
Our goal is to first observe the performance trends of algorithms according to our three classification criteria. Then, we will explore combinations that show particularly strong performance. 
\paragraph{Experiment environment}
We used RTX 3090 GPU, and Adam optimizer for all of trainings. 
We conducted training for each case 10 times, and the results are displayed as the mean (in the larger font above) and standard deviation (in the smaller font below) for both the main task metric (denoted by Perf) and constraint violation (denoted by Const.Vio)\footnote{For example, $\underset{\pm 00.77}{84.72}$ means that the average is 84.72, and the standard deviation is 00.77 from 10 experiments.} rate. 

\paragraph{\textbf{Metric}}
Comparing the superiority of experimental results considering two different metrics simultaneously is very challenging, especially when there is no occurrence of Pareto-improvement. The \textit{$H\beta$-score} (Harmonic $\beta$ Score) we propose is an indicator that allows for a quick and clear evaluation of experimental outcomes based on two metrics. Assume we have two metrics to consider, and denote the scores for each metric as $m_1$ and $m_2$, respectively. Both metrics are assumed to have values ranging from 0 to 1, with higher values indicating better performance\footnote{Constraint violation rate is used for table \ref{table:Experiment_results_total}. However, when we consider $H\beta$-score, we convert it to the constraint satisfaction rate, which is $1-$(constraint violation rate).}. The $H\beta$-score is similar in form to the F$\beta$-score and is defined as follow:

\[
H_{\beta}(m_1, m_2) = \frac{1+\beta^{2}}{\frac{1}{m_1} + \frac{\beta^2}{m_2}}.
\]
The $H\beta$-score is exactly the same in form as the F$\beta$-score. It is simply an extension of the F$\beta$-score, which uses precision and recall as arguments, to be a score for any two arbitrary metrics. If the magnitude of $\beta$ increases, the evaluation significantly considers the weight of $m_2$. Conversely, as the value of $\beta$ approaches zero, the weight of $m_1$ is significantly considered in the evaluation. 

\paragraph{\textbf{Experiment results}}
Table \ref{table:Experiment_results_total} shows the experiment results for all combinations possible in our analysis axes which consist of previous methods and our newly proposed methods. 
For each task, we present the experimental results based on our three analysis axes proposed in section \sref{sec:Analysis_Axes}: type of constraint loss (soft, binary, real),  exploration strategy of constraint-violating examples (top-1, sampling, exhaustive), and mechanism for integrating the main and the constraint information (static, monotone, proj-sup, proj-con, proj-both).

As the sheer number of experiments is too large to interpret in table \ref{table:Experiment_results_total}, we try to examine key factors for best main task performance, constraint violation by dissecting the table \ref{table:Experiment_results_total} from different perspectives. 

\paragraph{\textbf{Trends per analysis axes}}

Figure \ref{fig:tendency_integration_mechanism} illustrates 
the top 5 experimental results with the highest $H\beta$-scores 
for each of the five integration mechanisms described in section \sref{subsec:update}.
Among the five integration mechanisms, \textit{projection-con} and \textit{projection-both} consistently demonstrates the best performance across most $\beta$ values. They excel in a wide range of scenarios, from those emphasizing main task metrics (lower $\beta$ values) to those prioritizing constraint injection task performance (higher $\beta$ values). The static and monotone mechanism 
seldom 
performs well
, they do not always exhibit excellent performance across all tasks.

Figure \ref{fig:tendency_type_loss} illustrates 
the top 5 experimental results with the highest $H\beta$-scores for each of the three types of constraint losses described in section \sref{subsec:Constraint_Loss}.
Among the three types of losses, whether soft or real type shows consistently better performance depends on the task. Our hypothesis is, as mentioned in appendix \ref{fine_grained_expressiveness}, soft and real types of losses can incorporate more fine-grained information into constraint loss compared to binary types of loss.

Figure \ref{fig:tendency_exploring_strategy} illustrates 
the top 5 experimental results with the highest $H\beta$-scores, considering each of the five exploring strategies described in section \sref{subsec:exploration}.
Among the five strategies, one clear observation is that the sampling method consistently demonstrates superior performance across all tasks. Although there are variations, performance tends to improve as the sample size increases. However, the overall performance of the full strategy is not favorable, especially in SRL task. In the full strategy, the model generates errors that significantly different from those expected for realistic output, resulting in suboptimal performance due to the associated loss. 
We hypothesize that the full strategy’s performance of SRL is even worse than that observed in NLI, due to the significantly larger output space.

To summarize, sampling strategy and projection-con, projection-both mechanisms consistently demonstrate superior performance across all tasks. However, in relation to the type of constraint loss, there is no type of loss that consistently shows superior performance across all tasks; it varies depending on the task.
As shown in Table \ref{table:Experiment_results_total}, the number of combinations of learning algorithms with output constraints based on our analysis criteria is quite large (65 combinations for the NLI and SRL tasks, and 40 combinations for the STE task). Therefore, it is practically impossible to experiment with all learning algorithms. We examined the performance trends of the algorithms through figures 1, 2, and 3, which provide useful insights for selecting learning algorithms.

\begin{figure}[t]
  \includegraphics[width=\columnwidth]{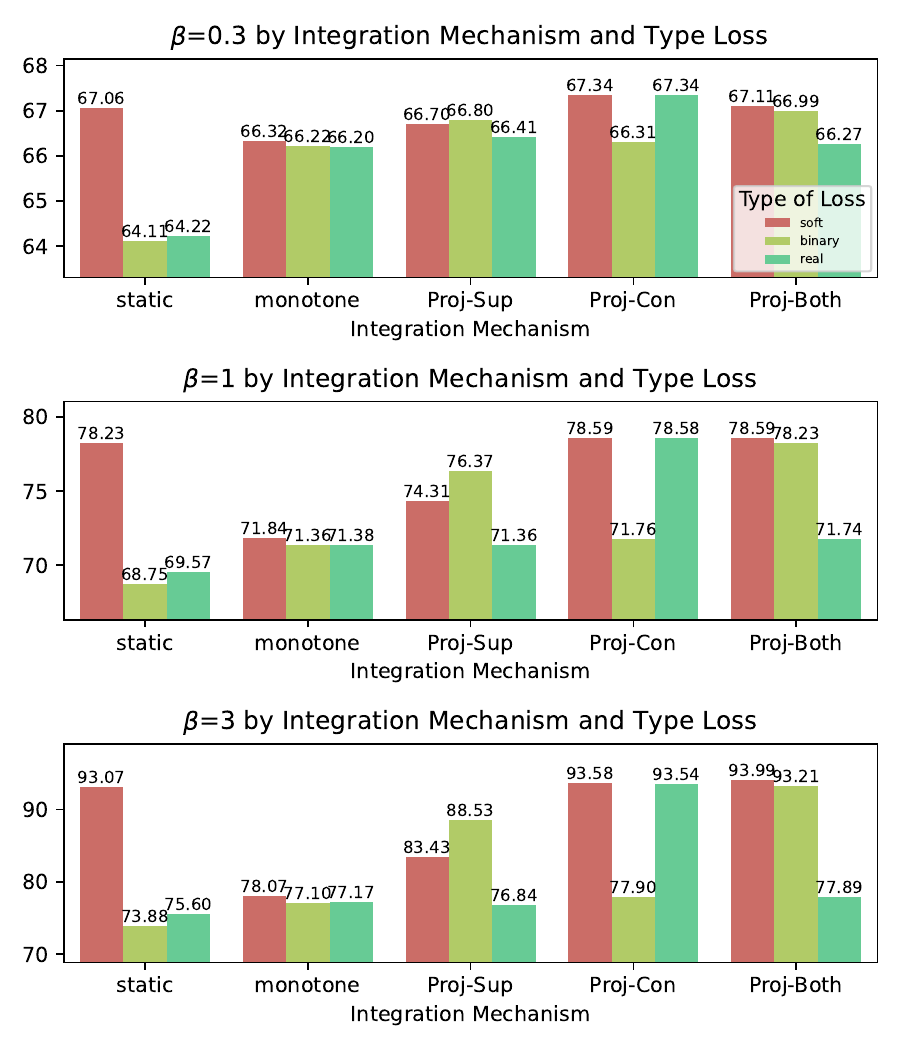}
  \caption{Experiment result from NLI task with samp-10. Three bar plots represents the $H\beta$-scores with respect to the integration mechanism (separated by the $x$-axis) and type of constraint losses (separated by the color). From top to bottom, the corresponding values of $\beta$'s are $0.3$, $1$, $3$, respectively.}
\label{fig:NLI_samp10_fixed}
\end{figure}

\begin{figure}[t]
  \includegraphics[width=\columnwidth]{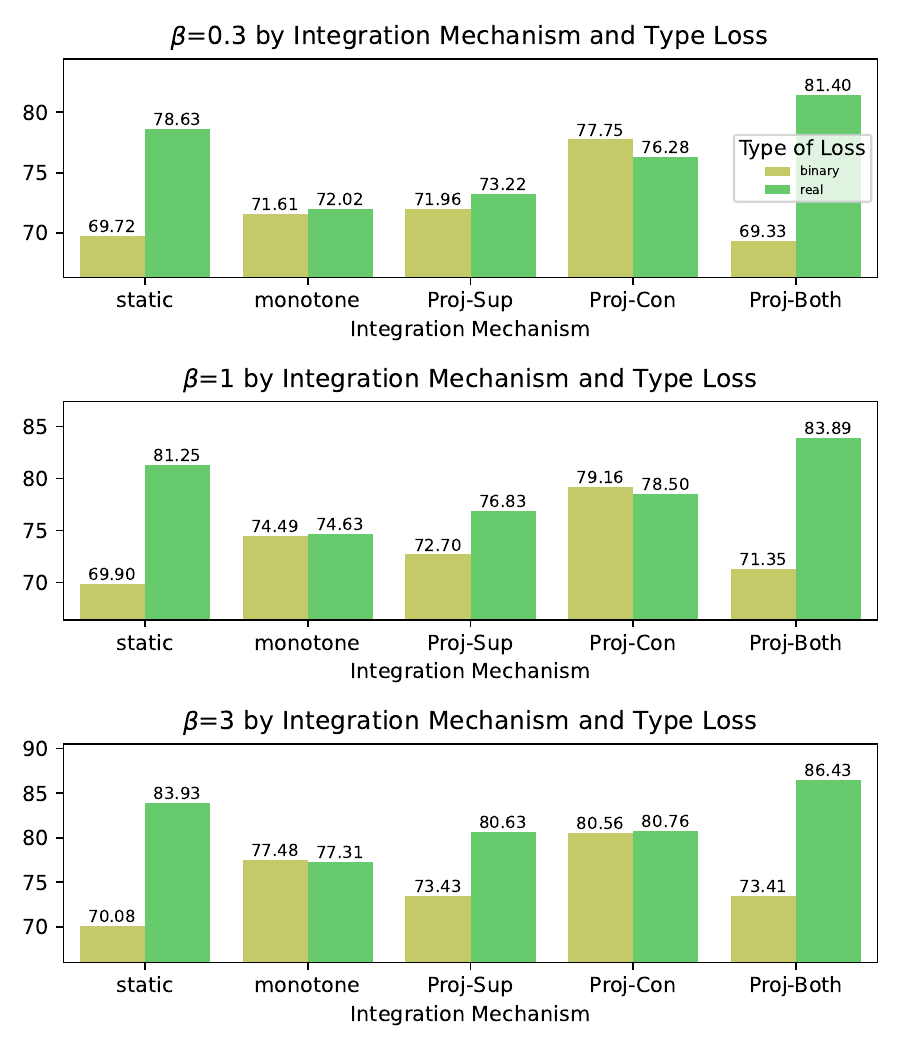}
  \caption{Experiment result from STE task with samp-10. Three bar plots represents the $H\beta$-scores with respect to the integration mechanism (separated by the $x$-axis) and type of constraint losses (separated by the color). From top to bottom, the corresponding values of $\beta$'s are $0.3$, $1$, $3$, respectively.}
\label{fig:STE_samp10_fixed}
\end{figure}

\begin{figure}[t]
  \includegraphics[width=\columnwidth]{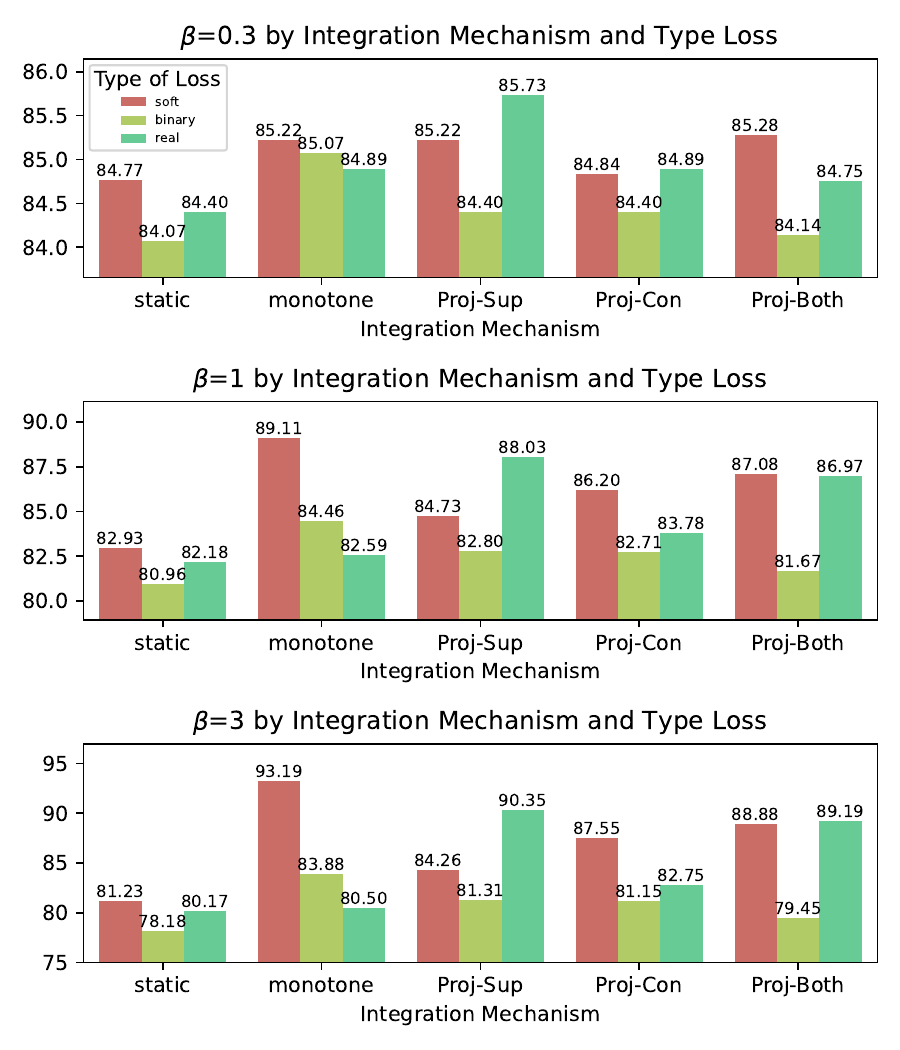}
  \caption{Experiment result from SRL task with samp-10. Three bar plots represents the $H\beta$-scores with respect to the integration mechanism (separated by the $x$-axis) and type of constraint losses (separated by the color). From top to bottom, the corresponding values of $\beta$'s are $0.3$, $1$, $3$, respectively.}
\label{fig:SRL_samp10_fixed}
\end{figure}

\paragraph{\textbf{Specific combinations of axes outperforming others}}
In addition to observing overall trends, we dive into a more detailed analysis of specific algorithm combinations and their performance.
We observed in the previous experimental results (figures \ref{fig:tendency_integration_mechanism}, \ref{fig:tendency_type_loss}, \ref{fig:tendency_exploring_strategy}) that the sampling strategy and projection-con, projection-both mechanisms generally perform well, with performance improving as sample size increases. However, the results in figures \ref{fig:tendency_integration_mechanism}, \ref{fig:tendency_type_loss}, and \ref{fig:tendency_exploring_strategy} represent averages across multiple algorithm outcomes and do not depict individual algorithms. In this section, we narrow our focus and present an analysis for individual algorithms assuming a fixed sampling strategy with a sample size of 10 (referred to as samp-10 from now on) which consistently performed the best across different tasks, across different conditions.
Figures \ref{fig:NLI_samp10_fixed}, \ref{fig:STE_samp10_fixed}, and \ref{fig:SRL_samp10_fixed} depict the $H\beta$-scores for different combinations of loss types and integration mechanisms when the sampling strategy is fixed as samp-10. For visibility, we consider  values of $0.3$, $1$, and $3$.

Notably, our newly proposed projection-based algorithms, projection-con and projection-both, exhibit the highest-level performance across most situations. One interesting point is the performance difference between projection-con and projection-both mechanisms. By examining the average of the top 5 number of $H\beta$-scores (as previously shown in figure \ref{fig:tendency_integration_mechanism}), we find that projection-con outperforms other mechanisms. 
However, upon observing individual algorithms per task, we found that for the soft or real types of loss, the projection-both mechanism shows the best-level performance than other mechanisms for most combinations.
In the case of the SRL task, there are instances where the monotone mechanism performs well. Particularly, when used in conjunction with a soft type of loss, the monotone mechanism exhibits higher performance, which is inconsistent with other experimental results. The reason for this discrepancy has not been clearly identified yet, but the specific characteristics of weight updates in constraint loss combined with a soft type of loss for achieving higher performance remain a subject for future work.


Another noteworthy observation is that under samp-10, the soft type of loss exhibits the highest performance in most cases. Results from figures \ref{fig:NLI_samp10_fixed} and \ref{fig:SRL_samp10_fixed} show that, except for the projection-sup instance in SRL, soft type of loss generally outperforms real type of loss. We previously observed from figure \ref{fig:tendency_type_loss} that real type of loss tends to perform best in the SRL task among the three types of losses. The results in \ref{fig:SRL_samp10_fixed}, however, demonstrate the opposite, indicating that soft type of loss performs exceptionally well when combined with the sampling strategy.

\section{Additional related work}
There are two stages where constraints can be injected: at inference time and learning time. 
At inference time, the goal is to remedy nonsensical outputs that violate human constraints at test time regardless of the training procedure
~\cite{lee2019gradient,roth2005integer}. 
For example, \cite{lee2019gradient} updates the model parameters at test time for each test instance to satisfy constraints, while \cite{roth2005integer} transforms the constrained problem into the form of integer linear programming for the inference process to maximize the log probability score with constraint satisfaction. 
Both methods have shown to satisfy constraints well and also improve performances, however, as these two methods utilize vastly different philosophies, their formulations are not directly comparable.

On the other hand, another line of research have explored how to practically use constraint injection in software development \cite{ahmed2022pylon,rajaby-faghihi-etal-2021-domiknows}, which demonstrate the application of constraint injection techniques in software. These tools can apply constraints across different domains, highlighting the versatility of constraint learning methods. Importantly, they are effective not only during the training phase but also during inference. This research and the platforms mentioned offer valuable insights into the practical application of constraints.

\section{Conclusions}
We have proposed three axes for classifying and categorizing learning algorithms related to injecting constraints: type of constraint loss, exploring strategy of constraint-violating examples, and integration of main task and constraint information. To the best of our knowledge, this study is the first to systematically classify existing learning algorithms with constraints under a unified formulation. We have analyzed the key factors that affect performance based on our analysis criteria, which helps in understanding learning algorithms with constraints.

Additionally, we have introduced three projection-based mechanisms as a novel approach for the integration mechanism of main task and constraint information. Viewing the main task and constraint injection as two separate tasks, we started with the motivation to prevent negative effects on each other during the gradient update process. This introduces a new perspective on integrating learning signals from main task and constraint, which shows superior performance compared to existing integration mechanisms.



\section{Limitations and future work}
Our experiments were exclusively conducted on NLP tasks (NLI, STE, SRL) and did not include cutting-edge large language models. Therefore, it would be worthwhile to extend our experiments to a broader range of tasks and larger models. This would not only validate the generalizability of our methods but also potentially uncover new insights and improvements for various applications.

\section*{Acknowledgments}
This work was supported in part by the National Research Foundation of Korea (NRF) grant (RS-2023-00280883, RS-2023-00222663), by the National Super computing Center with super computing resources including technical support (KSC-2023-CRE-0176), and partially supported by New Faculty Startup Fund from Seoul National University.

\bibliography{acl_latex}

\begin{thebibliography}{20}
\providecommand{\natexlab}[1]{#1}

\bibitem[{Ahmed et~al.(2022)Ahmed, Li, Ton, Guo, Chang, Kordjamshidi, Srikumar, Van~den Broeck, and Singh}]{ahmed2022pylon}
Kareem Ahmed, Tao Li, Thy Ton, Quan Guo, Kai-Wei Chang, Parisa Kordjamshidi, Vivek Srikumar, Guy Van~den Broeck, and Sameer Singh. 2022.
\newblock Pylon: A pytorch framework for learning with constraints.
\newblock In \emph{NeurIPS 2021 Competitions and Demonstrations Track}, pages 319--324. PMLR.

\bibitem[{Baczy{\'n}ski and Jayaram(2007)}]{baczynski2007characterizations}
Micha{\l} Baczy{\'n}ski and Balasubramaniam Jayaram. 2007.
\newblock On the characterizations of (s, n)-implications.
\newblock \emph{Fuzzy sets and systems}, 158(15):1713--1727.

\bibitem[{Bedregal et~al.(2010)Bedregal, Dimuro, Santiago, and Reiser}]{bedregal2010interval}
Benjam{\'\i}n~Callejas Bedregal, Gra{\c{c}}aliz~Pereira Dimuro, Regivan Hugo~Nunes Santiago, and Renata Hax~Sander Reiser. 2010.
\newblock On interval fuzzy s-implications.
\newblock \emph{Information Sciences}, 180(8):1373--1389.

\bibitem[{Bowman et~al.(2015)Bowman, Angeli, Potts, and Manning}]{bowman-etal-2015-large_snli}
Samuel~R. Bowman, Gabor Angeli, Christopher Potts, and Christopher~D. Manning. 2015.
\newblock \href {https://doi.org/10.18653/v1/D15-1075} {A large annotated corpus for learning natural language inference}.
\newblock In \emph{Proceedings of the 2015 Conference on Empirical Methods in Natural Language Processing}, pages 632--642, Lisbon, Portugal. Association for Computational Linguistics.

\bibitem[{Br\"{o}cheler et~al.(2010)Br\"{o}cheler, Mihalkova, and Getoor}]{brocheler2012probabilistic}
Matthias Br\"{o}cheler, Lilyana Mihalkova, and Lise Getoor. 2010.
\newblock Probabilistic similarity logic.
\newblock In \emph{Proceedings of the Twenty-Sixth Conference on Uncertainty in Artificial Intelligence}, UAI'10, page 73–82, Arlington, Virginia, USA. AUAI Press.

\bibitem[{Chaudhry et~al.(2018)Chaudhry, Ranzato, Rohrbach, and Elhoseiny}]{chaudhry2018efficient_agem}
Arslan Chaudhry, Marc'Aurelio Ranzato, Marcus Rohrbach, and Mohamed Elhoseiny. 2018.
\newblock Efficient lifelong learning with a-gem.
\newblock \emph{arXiv preprint arXiv:1812.00420}.

\bibitem[{Lee et~al.(2019)Lee, Mehta, Wick, Tristan, and Carbonell}]{lee2019gradient}
Jay~Yoon Lee, Sanket~Vaibhav Mehta, Michael Wick, Jean-Baptiste Tristan, and Jaime Carbonell. 2019.
\newblock Gradient-based inference for networks with output constraints.
\newblock In \emph{Proceedings of the AAAI Conference on Artificial Intelligence}, volume~33, pages 4147--4154.

\bibitem[{Li et~al.(2020)Li, Jawale, Palmer, and Srikumar}]{li-etal-2020-structured}
Tao Li, Parth~Anand Jawale, Martha Palmer, and Vivek Srikumar. 2020.
\newblock \href {https://doi.org/10.18653/v1/2020.acl-main.744} {Structured tuning for semantic role labeling}.
\newblock In \emph{Proceedings of the 58th Annual Meeting of the Association for Computational Linguistics}, pages 8402--8412, Online. Association for Computational Linguistics.

\bibitem[{Liu et~al.(2019)Liu, Ott, Goyal, Du, Joshi, Chen, Levy, Lewis, Zettlemoyer, and Stoyanov}]{liu2019roberta}
Yinhan Liu, Myle Ott, Naman Goyal, Jingfei Du, Mandar Joshi, Danqi Chen, Omer Levy, Mike Lewis, Luke Zettlemoyer, and Veselin Stoyanov. 2019.
\newblock Roberta: A robustly optimized bert pretraining approach.
\newblock \emph{arXiv preprint arXiv:1907.11692}.

\bibitem[{Lopez-Paz and Ranzato(2017)}]{lopez2017gradient}
David Lopez-Paz and Marc'Aurelio Ranzato. 2017.
\newblock Gradient episodic memory for continual learning.
\newblock \emph{Advances in neural information processing systems}, 30.

\bibitem[{Mehta et~al.(2018)Mehta, Lee, and Carbonell}]{mehta2018towards}
Sanket~Vaibhav Mehta, Jay~Yoon Lee, and Jaime Carbonell. 2018.
\newblock Towards semi-supervised learning for deep semantic role labeling.
\newblock \emph{arXiv preprint arXiv:1808.09543}.

\bibitem[{Minervini and Riedel(2018)}]{minervini-riedel-2018-adversarially}
Pasquale Minervini and Sebastian Riedel. 2018.
\newblock \href {https://doi.org/10.18653/v1/K18-1007} {Adversarially regularising neural {NLI} models to integrate logical background knowledge}.
\newblock In \emph{Proceedings of the 22nd Conference on Computational Natural Language Learning}, pages 65--74, Brussels, Belgium. Association for Computational Linguistics.

\bibitem[{Nandwani et~al.(2019)Nandwani, Pathak, and Singla}]{nandwani2019primal}
Yatin Nandwani, Abhishek Pathak, and Parag Singla. 2019.
\newblock A primal dual formulation for deep learning with constraints.
\newblock \emph{Advances in Neural Information Processing Systems}, 32.

\bibitem[{Pradhan et~al.(2012)Pradhan, Moschitti, Xue, Uryupina, and Zhang}]{pradhan2012conll}
Sameer Pradhan, Alessandro Moschitti, Nianwen Xue, Olga Uryupina, and Yuchen Zhang. 2012.
\newblock Conll-2012 shared task: Modeling multilingual unrestricted coreference in ontonotes.
\newblock In \emph{Joint conference on EMNLP and CoNLL-shared task}, pages 1--40.

\bibitem[{Rajaby~Faghihi et~al.(2021)Rajaby~Faghihi, Guo, Uszok, Nafar, and Kordjamshidi}]{rajaby-faghihi-etal-2021-domiknows}
Hossein Rajaby~Faghihi, Quan Guo, Andrzej Uszok, Aliakbar Nafar, and Parisa Kordjamshidi. 2021.
\newblock \href {https://doi.org/10.18653/v1/2021.emnlp-demo.27} {{D}omi{K}now{S}: A library for integration of symbolic domain knowledge in deep learning}.
\newblock In \emph{Proceedings of the 2021 Conference on Empirical Methods in Natural Language Processing: System Demonstrations}, pages 231--241, Online and Punta Cana, Dominican Republic. Association for Computational Linguistics.

\bibitem[{Rajaby~Faghihi et~al.(2023)Rajaby~Faghihi, Nafar, Zheng, Mirzaee, Zhang, Uszok, Wan, Premsri, Roth, and Kordjamshidi}]{gluecons}
Hossein Rajaby~Faghihi, Aliakbar Nafar, Chen Zheng, Roshanak Mirzaee, Yue Zhang, Andrzej Uszok, Alexander Wan, Tanawan Premsri, Dan Roth, and Parisa Kordjamshidi. 2023.
\newblock Gluecons: A generic benchmark for learning under constraints.
\newblock In \emph{Proceedings of the AAAI Conference on Artificial Intelligence}, volume~33, pages 9552--9561.

\bibitem[{Roth and Yih(2005)}]{roth2005integer}
Dan Roth and Wen-tau Yih. 2005.
\newblock Integer linear programming inference for conditional random fields.
\newblock In \emph{Proceedings of the 22nd international conference on Machine learning}, pages 736--743.

\bibitem[{Sutskever et~al.(2014)Sutskever, Vinyals, and Le}]{sutskever2014sequence}
Ilya Sutskever, Oriol Vinyals, and Quoc~V Le. 2014.
\newblock Sequence to sequence learning with neural networks.
\newblock \emph{Advances in neural information processing systems}, 27.

\bibitem[{Williams(1992)}]{williams1992simple}
Ronald~J Williams. 1992.
\newblock Simple statistical gradient-following algorithms for connectionist reinforcement learning.
\newblock \emph{Machine learning}, 8:229--256.

\bibitem[{Xu et~al.(2018)Xu, Zhang, Friedman, Liang, and Broeck}]{xu2018semantic}
Jingyi Xu, Zilu Zhang, Tal Friedman, Yitao Liang, and Guy Broeck. 2018.
\newblock A semantic loss function for deep learning with symbolic knowledge.
\newblock In \emph{International conference on machine learning}, pages 5502--5511. PMLR.

\end{thebibliography}

\appendix
\section{More specific comparison between types of constraint losses: \psl\ and REINFORCE}\label{appendix:psl_REINFORCE}

In this section, we dive deeper into the characteristics and applicability of the two types of constraint loss mentioned in Section 2.1: Probabilistic Soft Logic (\psl) and REINFORCE. These types of losses have unique strengths and weaknesses depending on the different circumstances.

\subsection{Fine-grained expressiveness of constraints}\label{fine_grained_expressiveness}
\psl\ stands superior in capturing more fine-grained information compared to REINFORCE. The \psl\ type of loss function evaluates not only the overall outcome but also performance of individual components to encourage more detailed feedback. For instance, consider a multi-label classification setting, where a particular book's every possible category needs to be predicted. 
An easily understandable example of constraint is associated with the hierarchical structure between labels: If a model predicts `science fiction', it must necessarily also make a prediction that includes a hierarchy higher than that, which is `fiction'. 
Note that,
the above hierarchical constraint can be considered in the form of conditional statement for propositions, as follows:
\begin{align}
\nonumber
    \text{Pred(science fiction)} \implies \text{Pred(fiction)}
\end{align}
For the sake of simplicity in explanation, we employ the Łukasiewicz logic for this example. 
The soft value corresponding to the above logical expression is:
\begin{align}\label{soft_value}
    \text{min}(1, 1 - P_{s} + P_{f})
\end{align}
, where $P_{s}$ and $P_{f}$ represent the probability of being predicted for the science fiction class, and fiction class, respectively. In constraint learning using \psl\, the learning process aims to increase the soft value \eqref{soft_value}. In this example, the learning is conducted to increase $P_f - P_s$.
Since `fiction' is a class with higher hierarchy, and more inclusive class than `science fiction', learning to increase $P_f - P_s$ is highly reasonable. Likewise, the \psl\ type of constraint loss can enrich the model's understanding and providing more detailed feedback.
In REINFORCE, however, if the model's prediction violates the constraint, the probability for the prediction is directly reflected in the loss, regardless of the constraint imposed. This makes it challenging to provide detailed information about specifically which part should we penalize in the model's prediction. To incorporate more fine-grained information in REINFORCE, there is research that utilizes real rewards. For example, \citeauthor{mehta2018towards} defines reward score as $s=1-2g \in [-1,1]$, where $g\in[0,1]$ stands for normalized error count, so that larger constraint violations lead to greater constraint loss. Although the loss function cannot reflect the soft value of logical expression, by assigning rewards differently based on the degree of constraint violation, it is possible to incorporate more fine-grained information into the loss function than just assigning binary rewards.
\begin{table}[t]
\resizebox{78.7mm}{13mm}{
\begin{tabular}{lccc}
\toprule
Logic&  Product & Gödel & Łukasiewicz \\ \midrule
Negation &$1-a$ & $1-a$&  $1-a$   \\
T-conorm &   $a+b-ab$  &  max$(a, b)$ &min$(1, a+b)$  \\
T-norm   &  $ab$  &  min$(a, b)$   &  max$(0, a+b-1)$ \\
Implication
& $\begin{cases}
    1 & \text{if } a \leq b\\
    b/a & \text{otherwise}
\end{cases}$
& $\begin{cases}
    1 & \text{if } a \leq b\\
    b & \text{otherwise}
\end{cases}$
& $\min(1, 1-a+b)$\\
 \bottomrule
\end{tabular}
}
\caption{Examples of logics. Our experiment used Gödel logic except for the implication ($\Rightarrow$). For $\Rightarrow$, we used S-implication \cite{baczynski2007characterizations,bedregal2010interval} form, $\max(1-a, b)$.}
\label{table:soft_logic_norms}
\end{table}

\subsection{Type of constraints that constraint loss can represent}\label{psl_reinforce_comparison}
Though \psl\ stands superior in capturing more fine-grained information compared to REINFORCE, \psl\ encounters difficulties when representing a variety of constraints, while REINFORCE can express arbitrary types of constraints.
\citeauthor{gluecons} introduced limitations in encoding a specific type of knowledge in research related to constraint injection during training time \cite{nandwani2019primal,ahmed2022pylon}. Instead of focusing on individual characteristics of these studies, we can generalize this limitation using our view of analysis axes. As in table \ref{table:limitation_to_encode}, we can rewrite the constraint types that each constraint loss type can handle, according to the two types of constraint loss: \psl\ and REINFORCE.

NC (Needs Conversion) in table \ref{table:limitation_to_encode} can sometimes be practically challenging due to significant overhead, making it difficult to leverage effectively. 
An example is seen in the STE task discussed in \sref{sec:selected_tasks}, where the constraint is defined as follows: the count of `b' in the target should be exactly three times that in the source. Let $s \in (az | bz)^{*}$ be an input sequence data, and $t$ be a predicted output sequence of model. Also, for a finite set $X=\{x_{1}, x_{2}, \cdots, x_{n}\} \subseteq \mathbb{N}$, let $s(X)$, $t(X)$ represent the presence of `b' at positions $x_1, ..., x_{n}$ in the sequences $s$ and $t$, respectively. 
Note that if the count of `b' in $s$ is 1, then the count of `b' in $t$ should be 3. While this represents a small portion of the original constraint, when expressed in the linearized logical expression for PSL application, it can be represented as follows:
\begin{align}
    \nonumber
    \overset{|s|}{\underset{i=1}{\bigwedge}}\left[
        s(\{i\}) \implies 
        \underset{1 \leq j_{1}< j_{2}< j_{3} \leq |t|}{
        \bigvee} t(\{j_{1}, j_{2}, j_{3}\})
    \right]
\end{align}
However, even this partial inclusion of the overall constraint requires an excessively high computational cost.
\begin{table}[t]
\resizebox{78.7mm}{10mm}{%
\begin{tabular}{lccccc}
\toprule
        & Seq  & Lin & Log & Log+Quan  & Prog \\ \midrule
\psl &\checkmark & \checkmark & NC &NC  & X   \\
REINFORCE  &\checkmark & \checkmark & \checkmark & \checkmark  & \checkmark  \\
 \bottomrule
\end{tabular}
}%
\caption{This table classifies constraints that constraint-injection methods can handle during training time. We reinterpret table 2 of \cite{gluecons} with our axes of analysis: type of constraint loss. The specific meaning of abbreviations are as follows: Seq=sequential structure, Lin=linear constraint, Log=logical constraint, Log+Quan=logical constraint with quantifier, Prog=any constraints encoded as a program, NC=needs conversion.}
\label{table:limitation_to_encode}
\end{table}
\newline

The comparison between two types of constriant losses illustrates that the appropriate type of loss may vary depending on task requirements and problem details, reflecting the inevitable trade-off between the level of detailed information about constraints and the scope of constraint representation.

\section{Experiment result}

\begin{figure*}[t]
  \includegraphics[width=\textwidth]{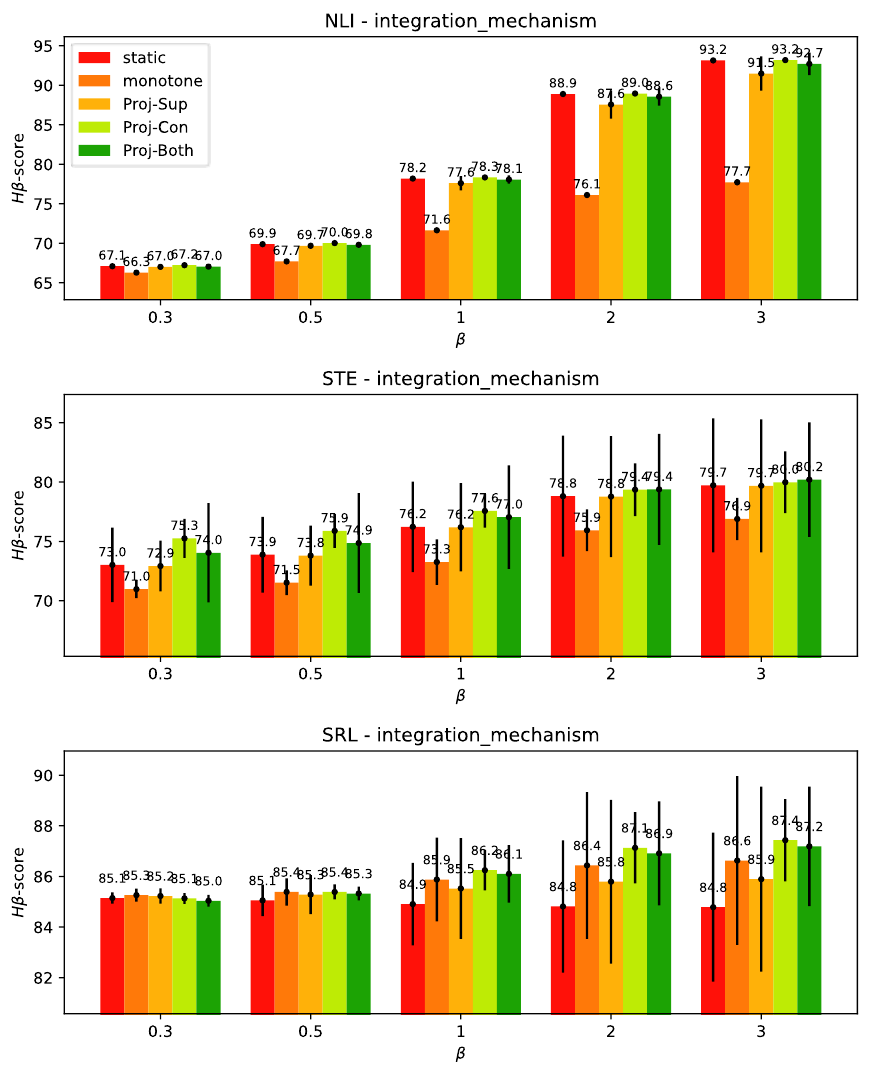}
  \caption{The $H\beta$-score values for different values of $\beta$ for three tasks: NLI, STE and SRL. For each $\beta$, the top 5 experimental results with the highest $H\beta$-scores are presented for each of the five integration mechanisms described in section \sref{subsec:update}.}
\label{fig:tendency_integration_mechanism}
\end{figure*}

\begin{figure*}[t]
  \includegraphics[width=\textwidth]{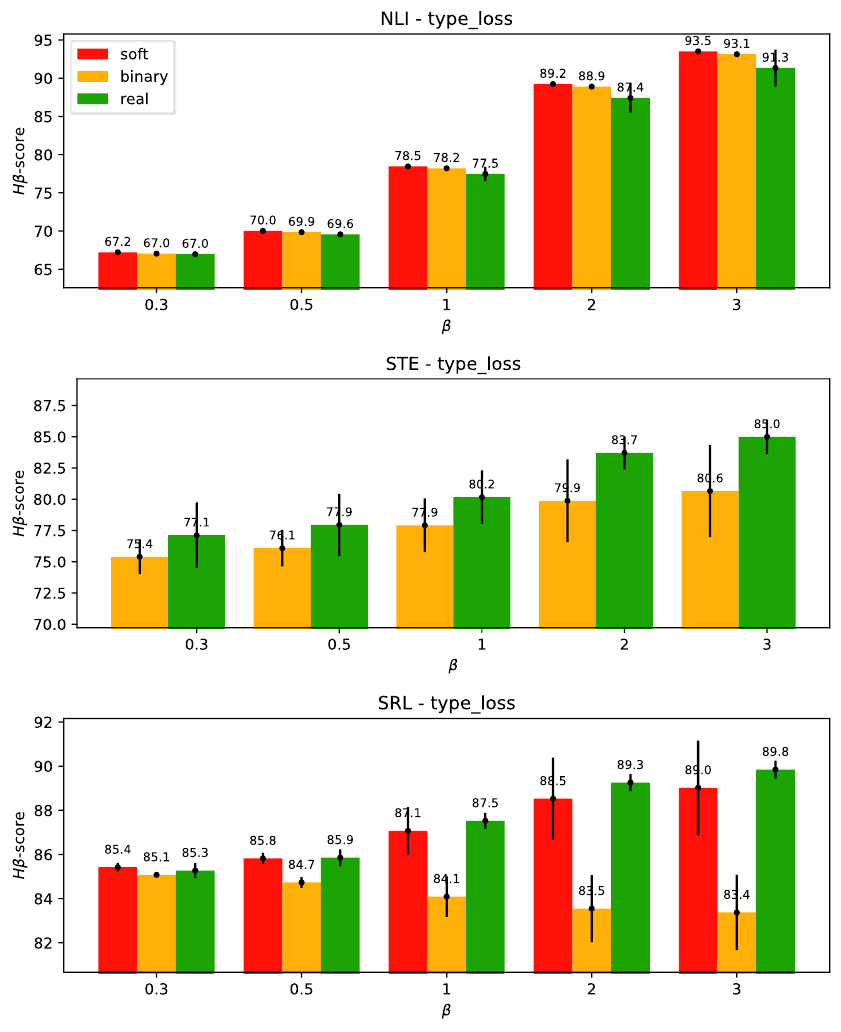}
  \caption{The $H\beta$-score values for different values of $\beta$ for three tasks: NLI, STE and SRL. For each $\beta$, the top 5 experimental results with the highest $H\beta$-scores are presented for each of the three types of constraint losses described in section \sref{subsec:Constraint_Loss}.}
\label{fig:tendency_type_loss}
\end{figure*}

\begin{figure*}[t]
  \includegraphics[width=\textwidth]{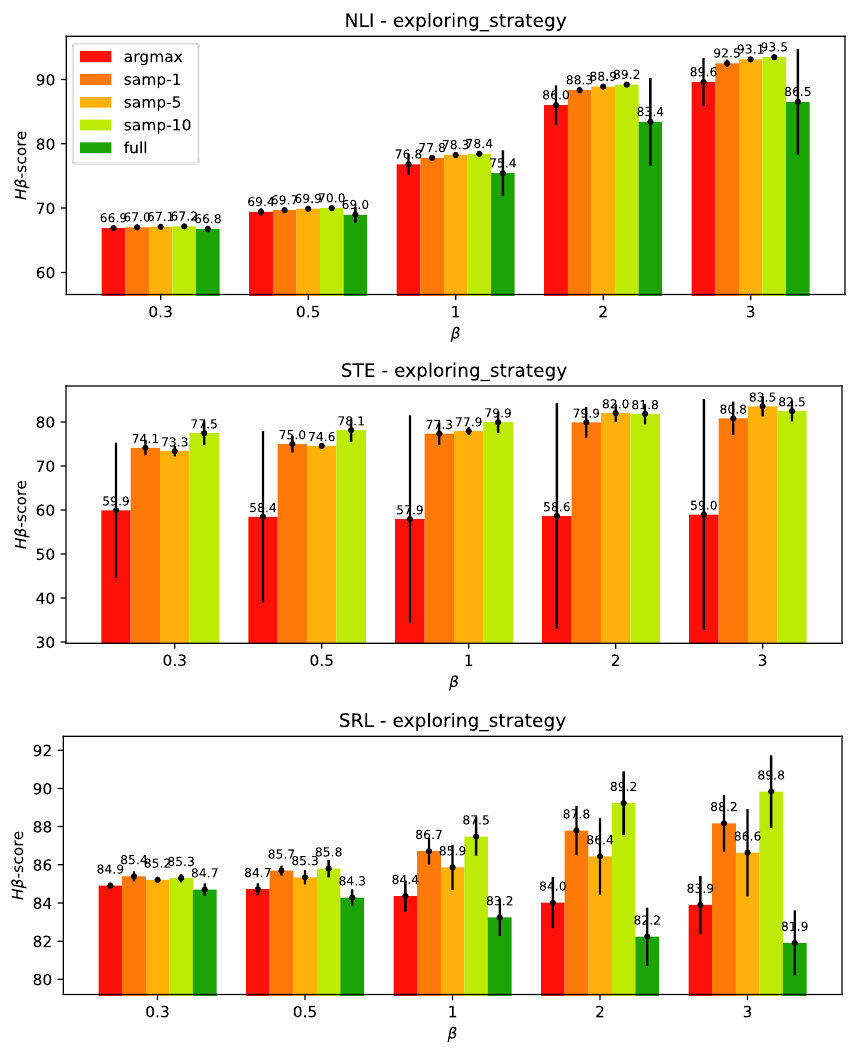}
  \caption{The $H\beta$-score values for different values of $\beta$ for three tasks: NLI, STE and SRL. For each $\beta$, the top 5 experimental results with the highest $H\beta$-scores are presented for each of the five integration mechanisms described in section \sref{subsec:exploration}.}
\label{fig:tendency_exploring_strategy}
\end{figure*}

Table \ref{table:Experiment_results_total} represents the experiment results for all combinations, containing main task metrics(\%, denoted as \say{Perf}) and constraint violation rates(\%, denoted as \say{Const.vio}). SRL, NLI, and STE tasks used F1 score, accuracy, and token accuracy for main task metric, respectively.  For the types of constraint losses, soft, binary, and real respectively represents \psl, REINFORCE method with binary reward, and REINFORCE method with real reward. 
The term `Baseline' refers to the experiment results without any constraint injection.
\citeauthor{ahmed2022pylon}, using the REINFORCE - binary reward method, separates the generation of constraint loss into two approaches in their implementation\footnote{https://github.com/pylon-lib/pylon}: one for decoded samples that satisfy the constraints and another for those that do not. In \citealp{mehta2018towards}, they generates constraint loss for decoded samples only when the constraints are violated. To compare various algorithms under a unified formulation, experiments involving constraint loss related to REINFORCE were conducted by generating constraint loss for examples that violated the constraints.

Note that we can easily extend the learning algorithms with constraints to semi-supervised learning. For SRL and NLI tasks, we also utilized unlabeled data during the training process. For SRL, we randomly selected $3\%$ of training data for unlabeled data. For NLI, we utilized the unlabeled data used in \cite{ahmed2022pylon}\footnote{https://github.com/pylon-lib/pylon/tree/master/examples/nli}.

\begin{table*}[h]  
\centering 
\scriptsize
\begin{adjustbox}{width=\textwidth}
\begin{tabular}{cllcccccccccc}
\toprule
Task&&   & \multicolumn{2}{c}{Top-1} & \multicolumn{2}{c}{Sampling-1}& \multicolumn{2}{c}{Sampling-5}& \multicolumn{2}{c}{Sampling-10} & \multicolumn{2}{c}{Exhaustive} \\
\midrule
&&   & Perf  & Const.Vio & Perf  & Const.Vio& Perf  & Const.Vio & Perf  & Const.Vio& Perf  & Const.Vio\\
\midrule

\multirow{29}{*}{NLI}&\multicolumn{2}{l}{Baseline}& \multicolumn{10}{c}{Acc: $\underset{\pm 00.30}{65.14}$, $\;$Const.Vio: $\underset{\pm 02.57}{20.81}$}\\
\cmidrule{2-13}

&\multirow{8}{*}{soft} & static &$\underset{\pm 00.35}{65.18}$& $\underset{\pm 00.88}{04.72}$ & $\underset{\pm 00.40}{65.40}$& $\underset{\pm 00.60}{03.60}$& $\underset{\pm 00.26}{65.31}$& $\underset{\pm 00.36}{02.23}$ &$\underset{\pm 00.40}{65.22}$& $\underset{\pm 01.88}{02.29}$  & $\underset{\pm 00.34}{65.20}$& $\underset{\pm 00.22}{01.95}$ \\[5.7pt]
&& monotone ($\lambda \uparrow$)&$\underset{\pm 00.26}{65.21}$& \textcolor{lightgray}{$\underset{\pm 03.03}{21.51}$}  &$\underset{\pm 00.27}{65.20}$& $\underset{\pm 02.09}{20.24}$ &$\underset{\pm 00.38}{65.30}$& \textcolor{lightgray}{$\underset{\pm 01.74}{21.83}$ }&$\underset{\pm 00.54}{65.33}$& $\underset{\pm 01.67}{20.20}$ &$\underset{\pm 00.55}{65.20}$ & \textcolor{lightgray}{$\underset{\pm 02.05}{22.72}$}\\[5.7pt]
&&Proj-Sup &$\underset{\pm 00.43}{65.28}$&$\underset{\pm 02.99}{20.14}$ &\textcolor{lightgray}{$\underset{\pm00.48}{65.05}$}&$\underset{\pm01.61}{20.73}$ &$\underset{\pm00.40}{65.26}$&$\underset{\pm00.38}{02.08}$ &$\underset{\pm00.20}{65.38}$&$\underset{\pm02.55}{13.93}$ &$\underset{\pm00.49}{65.36}$&$\underset{\pm00.41}{01.95}$ \\[5.7pt]
&&Proj-Con &$\underset{\pm00.27}{65.46}$&$\underset{\pm00.42}{03.05}$ &$\underset{\pm00.40}{65.23}$&$\underset{\pm00.74}{03.54}$ &\textcolor{lightgray}{$\underset{\pm00.26}{65.05}$}&$\underset{\pm00.33}{02.40}$ &\boldmath{$\underset{\pm00.28}{65.48}^{*}$}&$\underset{\pm00.17}{01.73}$ &$\underset{\pm00.29}{65.30}$&$\underset{\pm00.36}{02.50}$ \\[5.7pt]
&&Proj-Both &$\underset{\pm00.40}{65.20}$&$\underset{\pm03.13}{17.45}$ &$\underset{\pm00.29}{65.41}$&$\underset{\pm01.75}{06.16}$ &$\underset{\pm00.21}{65.39}$&$\underset{\pm05.04}{08.19}$ &$\underset{\pm00.32}{65.23}$&\boldmath{$\underset{\pm00.84}{01.17}^{*}$} &$\underset{\pm00.43}{65.40}$&\textcolor{lightgray}{$\underset{\pm03.29}{21.30}$} \\[5.7pt]
\cmidrule{2-13}
&\multirow{8}{*}{binary} & static &$\underset{\pm 00.38}{65.20}$&$\underset{\pm 02.65}{02.45}$  &\textcolor{lightgray}{$\underset{\pm 00.44}{64.23}$}& $\underset{\pm 05.29}{03.24}$ &\textcolor{lightgray}{$\underset{\pm 00.38}{63.26}$}& $\underset{\pm 11.67}{17.02}$ &\textcolor{lightgray}{$\underset{\pm 00.49}{63.26}$}& \textcolor{lightgray}{$\underset{\pm 02.40}{24.72}$}  & -&- \\[5.7pt]
&& monotone ($\lambda \uparrow$)&$\underset{\pm 00.31}{65.36}$& $\underset{\pm 03.38}{20.30}$ &\textcolor{lightgray}{$\underset{\pm 00.29}{65.10}$}& \textcolor{lightgray}{$\underset{\pm 02.66}{22.66}$ }&$\underset{\pm 00.37}{65.16}$& \textcolor{lightgray}{$\underset{\pm 01.78}{22.00}$} &$\underset{\pm 00.36}{65.29}$& \textcolor{lightgray}{$\underset{\pm 02.74}{21.32}$ }  & -&- \\[5.7pt]
&&Proj-Sup &$\underset{\pm00.35}{65.21}$&$\underset{\pm03.30}{14.73}$ &$\underset{\pm00.47}{65.25}$&$\underset{\pm01.66}{07.06}$ &$\underset{\pm00.23}{65.22}$&$\underset{\pm00.20}{02.25}$ &$\underset{\pm00.25}{65.18}$&$\underset{\pm04.97}{07.80}$ &-&- \\[5.7pt]
&&Proj-Con &\textcolor{lightgray}{$\underset{\pm00.49}{65.11}$}&$\underset{\pm06.08}{07.49}$ &\boldmath{$\underset{\pm00.19}{65.42}$}&$\underset{\pm02.43}{11.80}$ &$\underset{\pm00.35}{65.25}$&$\underset{\pm00.16}{02.57}$ &$\underset{\pm00.24}{65.33}$&$\underset{\pm03.01}{20.40}$ &-&- \\[5.7pt]
&&Proj-Both &$\underset{\pm00.28}{65.16}$&$\underset{\pm02.85}{19.76}$ &\textcolor{lightgray}{$\underset{\pm00.34}{65.05}$}&$\underset{\pm00.28}{02.11}$ &$\underset{\pm00.41}{65.23}$&\boldmath{$\underset{\pm00.37}{02.10}$} &$\underset{\pm00.49}{65.14}$&$\underset{\pm00.33}{02.18}$ &-&- \\[5.7pt]
\cmidrule{2-13}
&\multirow{8}{*}{real} & static &$\underset{\pm 00.24}{65.26}$ &$\underset{\pm 01.29}{20.60}$  &\textcolor{lightgray}{$\underset{\pm 00.34}{64.66}$}& $\underset{\pm 00.90}{01.92}$ &\textcolor{lightgray}{$\underset{\pm 00.44}{63.28}$}& \textcolor{lightgray}{$\underset{\pm 01.89}{23.82}$} &\textcolor{lightgray}{$\underset{\pm 00.30}{63.26}$}& \textcolor{lightgray}{$\underset{\pm 07.16}{22.72}$ }  & -&- \\[5.7pt]
&& monotone ($\lambda \uparrow$) &$\underset{\pm 00.21}{65.42}$& \textcolor{lightgray}{$\underset{\pm 01.88}{21.45}$} &$\underset{\pm 00.49}{65.14}$& \textcolor{lightgray}{$\underset{\pm 02.25}{20.97}$} &$\underset{\pm 00.38}{65.26}$& \textcolor{lightgray}{$\underset{\pm 02.26}{21.36}$} &$\underset{\pm 00.38}{65.26}$&\textcolor{lightgray}{ $\underset{\pm 03.68}{21.23}$} & -&- \\[5.7pt]
&&Proj-Sup &$\underset{\pm00.26}{65.26}$&$\underset{\pm01.96}{22.93}$ &$\underset{\pm00.25}{65.21}$&\textcolor{lightgray}{$\underset{\pm02.23}{22.70}$} &\textcolor{lightgray}{$\underset{\pm00.39}{65.11}$}&$\underset{\pm02.76}{20.68}$ &\boldmath{$\underset{\pm00.48}{65.51}$}&\textcolor{lightgray}{$\underset{\pm02.03}{21.65}$} &-&- \\[5.7pt]
&&Proj-Con &$\underset{\pm00.32}{65.27}$&\textcolor{lightgray}{$\underset{\pm01.67}{20.98}$} &$\underset{\pm00.36}{65.33}$&$\underset{\pm01.86}{08.75}$ &\textcolor{lightgray}{$\underset{\pm00.24}{65.04}$}&$\underset{\pm00.36}{02.39}$ &$\underset{\pm00.38}{65.49}$&\boldmath{$\underset{\pm00.20}{01.78}$} &-&- \\[5.7pt]
&&Proj-Both &\textcolor{lightgray}{$\underset{\pm00.34}{65.09}$}&\textcolor{lightgray}{$\underset{\pm01.27}{23.21}$} &$\underset{\pm00.31}{65.15}$&$\underset{\pm02.77}{07.04}$ &$\underset{\pm00.39}{65.40}$&$\underset{\pm03.42}{09.33}$ &$\underset{\pm00.26}{65.29}$&$\underset{\pm03.62}{20.40}$ &-&- \\[5.7pt]
\midrule\midrule

\multirow{20}{*}{STE}&\multicolumn{2}{l}{Baseline}& \multicolumn{10}{c}{Tok-Acc: $\underset{\pm 02.26}{67.26}$, $\;$Const.Vio: $\underset{\pm 10.07}{28.89}$}\\
\cmidrule{2-13}
&\multirow{8}{*}{binary} & static &$\underset{\pm 00.43}{69.98}$ & \textcolor{lightgray}{$\underset{\pm 12.11}{29.35}$}& $\underset{\pm 02.37}{73.59}$ & $\underset{\pm 12.72}{18.83}$&  $\underset{\pm 02.20}{69.35}$ & \textcolor{lightgray}{$\underset{\pm 06.73}{33.32}$}& $\underset{\pm 03.22}{69.68}$ & \textcolor{lightgray}{$\underset{\pm 17.58}{29.87}$}& -&- \\[5.7pt]
&& monotone ($\lambda \uparrow$)& \textcolor{lightgray}{$\underset{\pm 06.42}{67.17}$} & $\underset{\pm 13.09}{26.93}$& $\underset{\pm 05.72}{71.03}$ & \textcolor{lightgray}{$\underset{\pm 11.46}{33.55}$}& $\underset{\pm 03.79}{69.43}$ & $\underset{\pm 11.15}{24.89}$ & $\underset{\pm 04.63}{71.07}$ & $\underset{\pm 08.34}{21.74}$&-&-  \\[5.7pt]
&& Proj-Sup &\textcolor{lightgray}{$\underset{\pm 04.03}{43.55}$}&\textcolor{lightgray}{$\underset{\pm 03.65}{93.87}$} &$\underset{\pm 05.90}{75.01}$&\boldmath{$\underset{\pm 07.28}{10.86}$} &$\underset{\pm 03.67}{69.19}$&$\underset{\pm 16.22}{26.21}$ &$\underset{\pm 02.96}{71.81}$&$\underset{\pm 17.69}{26.38}$ &-&- \\[5.7pt]
&& Proj-Con &\textcolor{lightgray}{$\underset{\pm 03.27}{49.64}$}& \textcolor{lightgray}{$\underset{\pm 02.10}{98.40}$} &$\underset{\pm 03.97}{73.58}$&$\underset{\pm 16.16}{22.18}$ &$\underset{\pm 06.76}{74.96}$&$\underset{\pm 16.42}{22.71}$ &\boldmath{$\underset{\pm 08.92}{77.48}$}&$\underset{\pm 14.79}{19.08}$ &-&- \\[5.7pt]
&& Proj-Both &\textcolor{lightgray}{$\underset{\pm01.70}{51.19}$}&\textcolor{lightgray}{$\underset{\pm02.08}{98.77}$} &$\underset{\pm03.92}{70.71}$&$\underset{\pm12.72}{23.43}$ &$\underset{\pm03.34}{68.51}$&$\underset{\pm10.77}{26.62}$ &$\underset{\pm02.20}{68.94}$&$\underset{\pm15.75}{26.06}$ &-&- \\[5.7pt]
\cmidrule{2-13}
&\multirow{8}{*}{real} & static &  $\underset{\pm 04.25}{67.77}$ & \textcolor{lightgray}{$\underset{\pm 12.86}{30.18}$}&  $\underset{\pm 03.03}{70.16}$ & $\underset{\pm 12.04}{22.18}$& $\underset{\pm 04.02}{70.30}$ & \boldmath{$\underset{\pm 07.53}{10.86}^{*}$}& $\underset{\pm 05.00}{78.13}$ & $\underset{\pm 11.15}{15.37}$ &-&-  \\[5.7pt]
&& monotone ($\lambda \uparrow$) & \textcolor{lightgray}{$\underset{\pm 04.43}{63.73}$} & $\underset{\pm 13.45}{22.92}$ &\textcolor{lightgray}{$\underset{\pm 02.93}{60.74}$}& \textcolor{lightgray}{$\underset{\pm 34.51}{51.79}$}   &$\underset{\pm 03.29}{69.91}$& $\underset{\pm 13.41}{19.07}$ &$\underset{\pm 03.74}{71.53}$& $\underset{\pm 13.05}{21.99}$  &-&-\\[5.7pt]
&& Proj-Sup &\textcolor{lightgray}{$\underset{\pm 04.91}{46.30}$}&\textcolor{lightgray}{$\underset{\pm 04.92}{94.17}$} &\textcolor{lightgray}{$\underset{\pm 02.47}{54.32}$}&\textcolor{lightgray}{$\underset{\pm 01.72}{98.86}$} &$\underset{\pm 04.14}{73.02}$&$\underset{\pm 08.75}{15.04}$ &$\underset{\pm 02.49}{72.55}$&$\underset{\pm 10.61}{18.36}$ &-&- \\[5.7pt]
&& Proj-Con &\textcolor{lightgray}{$\underset{\pm 03.88}{48.20}$}&\textcolor{lightgray}{$\underset{\pm 04.58}{95.29}$} &\textcolor{lightgray}{$\underset{\pm 01.16}{52.98}$}&\textcolor{lightgray}{$\underset{\pm 00.73}{99.74}$} &$\underset{\pm 03.04}{72.24}$&$\underset{\pm 07.57}{14.26}$ &$\underset{\pm 03.16}{75.86}$&$\underset{\pm 08.36}{18.66}$ &-&- \\[5.7pt]
&& Proj-Both &\textcolor{lightgray}{$\underset{\pm02.60}{50.60}$}&\textcolor{lightgray}{$\underset{\pm03.16}{98.68}$} &$\underset{\pm05.59}{74.81}$&$\underset{\pm16.24}{17.40}$ &$\underset{\pm04.31}{72.00}$&$\underset{\pm14.56}{15.03}$ &\boldmath{$\underset{\pm04.24}{80.92}^{*}$}&$\underset{\pm06.07}{12.91}$ &-&- \\[5.7pt]
\midrule
\midrule

\multirow{29}{*}{SRL}&\multicolumn{2}{l}{Baseline}& \multicolumn{10}{c}{F1: $\underset{\pm 00.77}{84.72}$, $\;$Const.Vio: $\underset{\pm 04.09}{20.43}$}\\
\cmidrule{2-13}
&\multirow{8}{*}{soft} &static &$\underset{\pm01.49}{85.24}$&$\underset{\pm02.04}{15.17}$ &$\underset{\pm01.55}{85.02}$&$\underset{\pm03.02}{14.07}$ &$\underset{\pm01.13}{85.21}$&$\underset{\pm02.80}{19.53}$ &$\underset{\pm00.74}{85.15}$&$\underset{\pm36.95}{19.18}$ &$\underset{\pm00.98}{85.31}$&\textcolor{lightgray}{$\underset{\pm04.61}{21.72}$} \\[5.7pt]
&& monotone ($\lambda \uparrow$) &\textcolor{lightgray}{$\underset{\pm01.07}{84.42}$}&$\underset{\pm04.44}{18.53}$ &\boldmath{$\underset{\pm01.46}{85.78}^{*}$}&$\underset{\pm03.66}{14.40}$ &$\underset{\pm01.23}{85.12}$&$\underset{\pm03.12}{16.38}$ &\textcolor{lightgray}{$\underset{\pm01.16}{84.49}$}&\boldmath{$\underset{\pm01.42}{05.73}^{*}$} &$\underset{\pm00.81}{85.18}$&$\underset{\pm02.97}{15.97}$ \\[5.7pt]

&& Proj-Sup &$\underset{\pm00.98}{85.02}$&\textcolor{lightgray}{$\underset{\pm04.63}{20.79}$} &$\underset{\pm01.24}{85.19}$&$\underset{\pm04.85}{20.23}$ &$\underset{\pm01.03}{85.09}$&$\underset{\pm03.93}{19.59}$ &$\underset{\pm01.26}{85.32}$&$\underset{\pm04.08}{15.86}$ &$\underset{\pm00.97}{85.18}$&$\underset{\pm04.03}{18.73}$ \\[5.7pt]
&& Proj-Con &$\underset{\pm00.60}{84.96}$&$\underset{\pm01.23}{15.72}$ &$\underset{\pm01.53}{85.24}$&$\underset{\pm02.44}{11.76}$ &$\underset{\pm00.90}{85.07}$&$\underset{\pm02.85}{12.80}$ &\textcolor{lightgray}{$\underset{\pm01.17}{84.58}$}&$\underset{\pm03.61}{12.11}$ &\textcolor{lightgray}{$\underset{\pm00.57}{84.31}$}&$\underset{\pm04.01}{18.04}$ \\[5.7pt]
&& Proj-Both &$\underset{\pm01.21}{85.21}$&\textcolor{lightgray}{$\underset{\pm03.13}{21.86}$} &\textcolor{lightgray}{$\underset{\pm01.06}{84.71}$}&$\underset{\pm00.37}{12.11}$ &$\underset{\pm01.68}{85.62}$&$\underset{\pm03.57}{17.68}$ &$\underset{\pm02.06}{84.93}$&$\underset{\pm01.83}{10.66}$ &$\underset{\pm01.11}{85.03}$&$\underset{\pm04.26}{17.64}$ \\[5.7pt]
\cmidrule{2-13}
&\multirow{8}{*}{binary} & static &$\underset{\pm01.51}{85.20}$&$\underset{\pm00.66}{19.84}$ &$\underset{\pm01.02}{85.52}$&$\underset{\pm02.34}{19.53}$ &$\underset{\pm00.91}{85.19}$&$\underset{\pm02.91}{18.90}$ &\textcolor{lightgray}{$\underset{\pm01.28}{84.71}$}&\textcolor{lightgray}{$\underset{\pm04.34}{22.48}$} &-&- \\[5.7pt]
&& monotone ($\lambda \uparrow$) &\textcolor{lightgray}{$\underset{\pm00.94}{84.51}$}&\boldmath{$\underset{\pm02.93}{14.25}$} &\textcolor{lightgray}{$\underset{\pm00.75}{84.36}$}&\textcolor{lightgray}{$\underset{\pm05.88}{20.98}$ }&$\underset{\pm01.44}{85.23}$&$\underset{\pm03.37}{15.50}$ &$\underset{\pm00.68}{85.19}$&$\underset{\pm04.40}{16.26}$ &-&- \\[5.7pt]
&& Proj-Sup &\textcolor{lightgray}{$\underset{\pm01.05}{84.14}$}&\textcolor{lightgray}{$\underset{\pm02.12}{21.20}$} &\textcolor{lightgray}{$\underset{\pm01.20}{84.57}$}&$\underset{\pm02.20}{19.99}$ &\boldmath{$\underset{\pm01.01}{85.70}$}&\textcolor{lightgray}{$\underset{\pm03.92}{22.44}$} &$\underset{\pm00.57}{84.73}$&$\underset{\pm03.03}{19.05}$ &-&- \\[5.7pt]
&& Proj-Con &\textcolor{lightgray}{$\underset{\pm00.71}{84.54}$}&\textcolor{lightgray}{$\underset{\pm03.91}{21.28}$} &$\underset{\pm01.23}{85.56}$&$\underset{\pm04.38}{19.62}$ &$\underset{\pm01.03}{85.06}$&$\underset{\pm03.58}{19.79}$ &$\underset{\pm00.86}{84.74}$&$\underset{\pm06.10}{19.23}$ &-&- \\[5.7pt]
&& Proj-Both &$\underset{\pm00.77}{85.23}$&$\underset{\pm04.22}{20.36}$ &$\underset{\pm01.36}{84.89}$&$\underset{\pm03.33}{19.14}$ &$\underset{\pm01.014}{84.85}$&$\underset{\pm03.91}{19.46}$ &\textcolor{lightgray}{$\underset{\pm00.63}{84.61}$}&\textcolor{lightgray}{$\underset{\pm03.39}{21.09}$} &-&- \\[5.7pt]

\cmidrule{2-13}
&\multirow{8}{*}{real} & static &$\underset{\pm00.68}{85.05}$&$\underset{\pm03.21}{18.26}$ &$\underset{\pm00.96}{85.12}$&$\underset{\pm03.15}{09.79}$ &$\underset{\pm01.09}{85.33}$&\textcolor{lightgray}{$\underset{\pm04.43}{22.37}$} &$\underset{\pm01.23}{84.85}$&$\underset{\pm04.10}{20.32}$ && \\[5.7pt]
&& monotone ($\lambda \uparrow$) &$\underset{\pm00.78}{84.73}$&\textcolor{lightgray}{$\underset{\pm04.21}{21.99}$} &$\underset{\pm00.96}{85.09}$&\textcolor{lightgray}{$\underset{\pm03.95}{23.78}$} &$\underset{\pm01.29}{84.94}$&$\underset{\pm04.50}{19.61}$ &\boldmath{$\underset{\pm01.14}{85.36}$}&$\underset{\pm02.65}{20.00}$ &-&- \\[5.7pt]
&& Proj-Sup &$\underset{\pm01.46}{85.19}$&$\underset{\pm04.13}{19.19}$ &$\underset{\pm00.90}{85.09}$&$\underset{\pm04.40}{17.16}$ &$\underset{\pm00.94}{84.87}$&$\underset{\pm02.60}{09.21}$ &$\underset{\pm01.25}{85.29}$&\boldmath{$\underset{\pm02.20}{09.05}$} &-&- \\[5.7pt]
&& Proj-Con &$\underset{\pm00.93}{85.06}$&$\underset{\pm04.32}{18.22}$ &\textcolor{lightgray}{$\underset{\pm01.67}{84.31}$}&$\underset{\pm04.19}{09.47}$ &$\underset{\pm01.32}{85.19}$&\textcolor{lightgray}{$\underset{\pm05.09}{22.55}$} &$\underset{\pm01.08}{85.11}$&$\underset{\pm03.98}{17.50}$ &-&- \\[5.7pt]
&& Proj-Both &$\underset{\pm00.94}{85.05}$&$\underset{\pm04.55}{18.54}$ &$\underset{\pm00.74}{85.25}$&$\underset{\pm04.28}{20.36}$ &\textcolor{lightgray}{$\underset{\pm00.41}{84.54}$}&$\underset{\pm02.87}{11.96}$ &\textcolor{lightgray}{$\underset{\pm07.19}{84.33}$}&$\underset{\pm03.35}{10.23}$ &-&-\\[5.7pt]

\bottomrule
\end{tabular}
\end{adjustbox}
\caption{Experiment results for all combinations. The gray-colored numbers represent results with main task metrics and constraint violations worse than the baseline. For each type of constraint loss, results showing the highest main task metric and lowest constraint violation are highlighted in bold. For individual task, the highest main task metric and lowest constraint violation results are marked with an asterisk (*). In SRL and STE tasks, where the output takes the form of more than one token, the method of selecting the class with the highest probability for each token was employed for Top-1 strategy.}
\label{table:Experiment_results_total}
\end{table*}

\section{Pseudo-code for projection based integration mechanism}
\label{sec:appendix_PCCW_method}
Algorithm \ref{algo:PCCW} shows the detail pseudo-code for \textit{projection-both} mechanism. Pseudo-codes for projection-sup and projection-con mechanisms are variant of algorithm \ref{algo:PCCW}. For projection-sup, there is no need to store $g_{con}^{ref}$, nor to calculate the dot product between $g_{sup}$ and $g_{con}^{ref}$. Likewise, for projection-con, there is no need to store $g_{sup}^{ref}$, nor to calculate the dot product between $g_{con}$ and $g_{sup}^{ref}$.
\begin{algorithm}[tb]
\caption{Pseudo code for projection-both mechanism}
\label{algo:PCCW}
\textbf{Input}: labeled data $\mathcal{D}_{L} = \langle x_{i}, y_{i} \rangle_{i=1}^{T}$, unlabeled data $\mathcal{D}_{U} = \langle x_{i}^{u} \rangle_{i=1}^{T}$ (if available), model parameter $\theta$.
\begin{algorithmic}[1]
\STATE Initialize: $g_{sup}^{ref}\gets 0$, $g_{con}^{ref} \gets 0$.
  \WHILE{not converge}
  \STATE $\langle x_{L}, y_{L} \rangle \gets \text{sample from }\mathcal{D}_{L}$
  \STATE $\langle x_{U} \rangle \gets \text{sample from }\mathcal{D}_{U}$
  \STATE $g_{sup} \gets \nabla\mathcal{L}(x_L, y_L; \theta)$
  \STATE $g_{con} \gets \nabla(\mathcal{C}(x_L ; \theta)+\mathcal{C}(x_U ; \theta))$
  \IF{$g_{sup} \cdot g_{con}^{ref} < 0$}
    \STATE $g_{sup} \gets \text{ project }g_{sup} \text{ via } g_{con}^{ref}$
  \ENDIF
  \IF{$g_{con} \cdot g_{sup}^{ref} < 0$}
    \STATE $g_{con} \gets \text{ project }g_{con} \text{ via } g_{sup}^{ref}$
  \ENDIF
  \STATE $g_{sup}^{ref} \gets$ store the averaged vector of $g_{sup}$ across gradient updates.
  \STATE $g_{con}^{ref} \gets$ store the averaged vector of $g_{con}$ across gradient updates.

  \STATE Gradient update of $\theta$ for the cumulative gradients: $g_{sup}$ and $g_{con}$. 
  \ENDWHILE
\end{algorithmic}
\end{algorithm}

\section{Additional details about selected tasks and implementations}\label{sec:selected_tasks_app}
\subsection{SRL}
The baseline employs the RoBERTa baseline model \cite{liu2019roberta}, and two linear layers are added after the last layer of RoBERTa. While the parameters of the RoBERTa model are fixed, only the parameters of the last two linear layers are trained.

The model predicts one of 9 tags -\( \{ \texttt{O, B0, I0, ..., B3, I3}\}\) - and transforms all other tags into \texttt{O}. During the training process, 3\% of the data is randomly sampled from the training data for use.

For the real type of constraint loss in REINFORCE algorithm, the method employed to assign rewards is based on the count of duplicates in \texttt{B}. For all types of \texttt{B-X} that appear more than once, we summed the occurrences of all number of constraint-violated \texttt{B-X} and divided by the total sequence length, and this is multiplied by the constraint loss.

\subsection{NLI}
\begin{table}[t]
\resizebox{78.7mm}{16mm}{%
\begin{tabular}{cl}
\toprule
\multicolumn{2}{c}{\textbf{NLI Rules}}\\
\midrule
R1 &  $\text{T} \implies \text{ent}( X_{1}, X_{1})$ \\
R2 &  $\text{con}( X_{1}, X_{2}) \implies \text{con}( X_{2}, X_{1})$ \\
R3   &  $\text{ent}( X_{1}, X_{2}) \implies \neg\text{con}( X_{2}, X_{1})$ \\
R4 &  $\text{neu}( X_{1}, X_{2}) \implies \neg\text{con}( X_{2}, X_{1})$ \\
R5 & $\text{ent}( X_{1}, X_{2}) \wedge \text{ent}( X_{2}, X_{3}) \implies \text{ent}( X_{1}, X_{1})$\\
\bottomrule
\end{tabular}
}%
\caption{NLI Rules in \cite{minervini-riedel-2018-adversarially}.}
\label{table:NLI Rules}
\end{table}
The baseline employs the RoBERTa baseline model \cite{liu2019roberta}, and two linear layers are added after the last layer of RoBERTa. While the parameters of the RoBERTa model are fixed, only the parameters of the last two linear layers are trained. During training, 20\% of the training data is randomly sampled for use.

For the real type of constraint loss in REINFORCE algorithm, the method employed to assign rewards is based on the value in the \psl. In cases where it violates constraints, the corresponding \psl\ values are multiplied by the constraint loss.

\subsection{STE}
The training data includes 3 to 6 instances of az' and bz' in the source language, generating a dataset of 6000 instances. The test data comprises 3 to 8 instances of `az' and `bz' in the source language, transformed into the target language. We utilize seq2seq \cite{sutskever2014sequence} LSTM for prediction.

For the real type of constraint loss in REINFORCE algorithm, the method employed to assign rewards is a length-normalized quadratic: $(3x_{b}-y_{b})^{2}/(\text{len}(x) + \text{len}(y))$, where $x$ and $y$ respectively represents the input and output, while $x_{b}$ and $y_{b}$ respectively represents the number of occurrences of `b' in the input and output. 

We don't utilize a \psl\ type of constraint loss for STE task. This is because expressing constraints about the number of `b' occurrences in input and output is highly intricate for \psl. This constraint serves as an example demonstrating the difficulty of applying \psl\ to all types of constraints.

\end{document}